\documentclass[conference]{IEEEtran}
\IEEEoverridecommandlockouts
\usepackage{cite}
\usepackage{amsmath,amssymb,amsfonts}
\usepackage{algorithmic}
\usepackage{graphicx}
\usepackage{textcomp}
\usepackage{xcolor}
\usepackage{hyperref}
\usepackage{pbox}
\usepackage{caption}
\usepackage{subfig}

\def\BibTeX{{\rm B\kern-.05em{\sc i\kern-.025em b}\kern-.08em
    T\kern-.1667em\lower.7ex\hbox{E}\kern-.125emX}}
\begin{document}

\title{Autocamera Calibration for traffic surveillance cameras with wide angle lenses.\\
}

\author{\IEEEauthorblockN{Aman Jain}
\IEEEauthorblockA{\textit{Department of Mechanical Engineering} \\
\textit{Visvesvaraya National Institute of Technology}\\
Nagpur, India \\
amanjain@students.vnit.ac.in}
\and
\IEEEauthorblockN{Nicolas Saunier}
\IEEEauthorblockA{\textit{Department of Civil, Geological and Mining Engineering} \\
\textit{Polytechnique Montr\'eal}\\
Montreal, Canada \\
nicolas.saunier@polymtl.ca}

}

\maketitle

\begin{abstract}
 We propose a method for automatic calibration of a traffic surveillance camera  with wide-angle lenses. Video footage of a few minutes is sufficient for the entire calibration process to take place. This method takes in the height of the camera from the ground plane as the only user input to overcome the scale ambiguity. The calibration is performed in two stages, 1. Intrinsic Calibration 2. Extrinsic Calibration. Intrinsic calibration is achieved by assuming an equidistant fish-eye distortion and an ideal camera model. Extrinsic calibration is accomplished by estimating the two vanishing points, on the ground plane, from the motion of vehicles at perpendicular intersections. The first stage of intrinsic calibration is also valid for thermal cameras. Experiments have been conducted to demonstrate the effectiveness of this approach on visible as well as thermal cameras.

\end{abstract}

\begin{IEEEkeywords}
fish-eye, calibration, thermal camera, intelligent transportation systems, vanishing points
\end{IEEEkeywords}

\section{Introduction}
Camera calibration is of immense importance in the extraction of information from video surveillance data. It could either be used to deal with the perspective distortion of the object in the image plane or it can be used for photogrammetric measurements like distances, velocities, trajectories, etc. It is also fundamental for performing the multiview 3D reconstruction. Besides, with the aid of 3D information, it could also be used for vehicle tracking or object detection, robust to occlusion.

Owing to its importance, a significant portion of literature in computer vision addresses the problem of camera calibration. Almost all the methods of calibration could be categorized into two major approaches:-
\begin{itemize}
    \item Finding image-world correspondences\cite{MITpress, Hartley:2003:MVG:861369,Tsai, Zhang}.
    \item Vanishing Point-based methods\cite{Caprile1990, 10.1007/3-540-47979-1_12, Zisserman, 1661552, 4587780}.
\end{itemize} The first method tries to exploits the properties of the 3D scene structure to find correspondences between the real-world and its 2D image captured by the camera. With the aid of these correspondences, intrinsic as well as extrinsic estimation could be made. Further, the accuracy can be improved with increasing such correspondences. The vanishing point methods are majorly based on estimating the orthogonal vanishing points in the scene and requires no a priori knowledge for recovering the extrinsic and intrinsics matrices.

If the camera is already pre-calibrated using 3D rigs or checker-board, this data could be used directly in the second stage. However, in most cases, this data is not always available and hence there is a need for automatic intrinsic calibration also. Most of the literature in traffic surveillance considers an ideal camera model which assumes that the pixels are perfectly square with zero skew and the optical center of the camera coincides exactly with the image center. The last assumption is not necessarily true and in such cases, the optical center is calculated in a slightly different fashion.  

For extrinsic calibration in the context of traffic surveillance, the correspondences approach may require annotation of lane marking with its lane-width\cite{article}\cite{doi:10.1139/cjce-2011-0456} or presence of ground control points, or using regional heuristics such as average vehicle dimensions\cite{bhardwaj2017autocalib} or speed. However, the majority of the above methods involve a human intervention or are location-dependent and are unsuitable for generalized auto-calibration. Thus for such applications, we make use of vanishing points. The vanishing points may be generated from the static scene structures or lane markings, or motion of vehicles and pedestrians\cite{4020751}. For robust auto camera calibration, it is always advisable that the calibration process is independent of the scene in general and hence our approach will use only the moving objects in the scene.
This paper especially deals with the wide-angle lenses cameras, that are ideal for most of the photogrammetry applications. However, they are always accompanied by various distortion effects, among which fisheye  effects are dominating. It is essential to remove such effects before vanishing points estimation. The remaining paper is organized into the following sections: Section II defines the Camera Model, Section III and IV describes the process of Intrinsic as well as Extrinsic Calibration, Section V presents the results of our approach and Section VI explains the Conclusion and Future Scope of the algorithm.
\begin{figure*}
\begin{tabular}{cc}
    \centering
    \subfloat[]{\includegraphics[width=8cm]{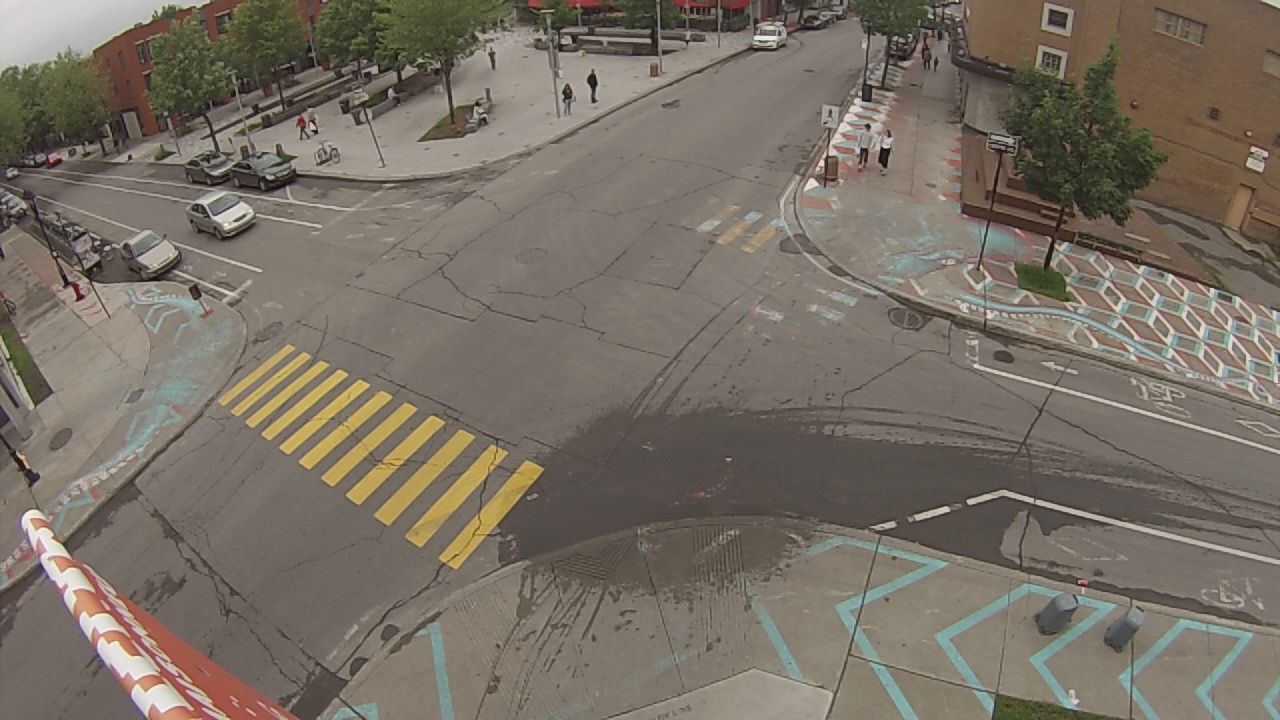}}&
    \subfloat[]{\includegraphics[width=8cm]{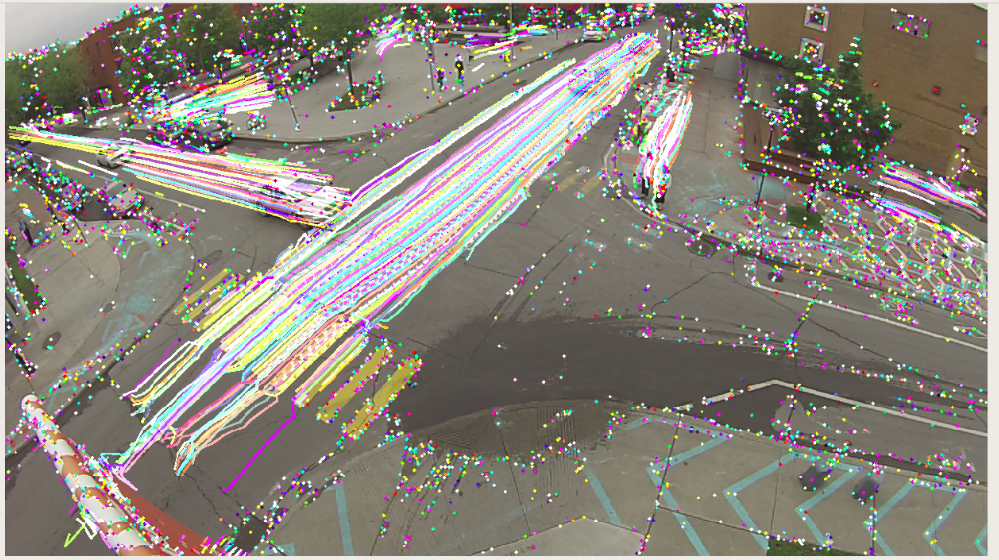}}\\
    \subfloat[]{\includegraphics[width=8cm]{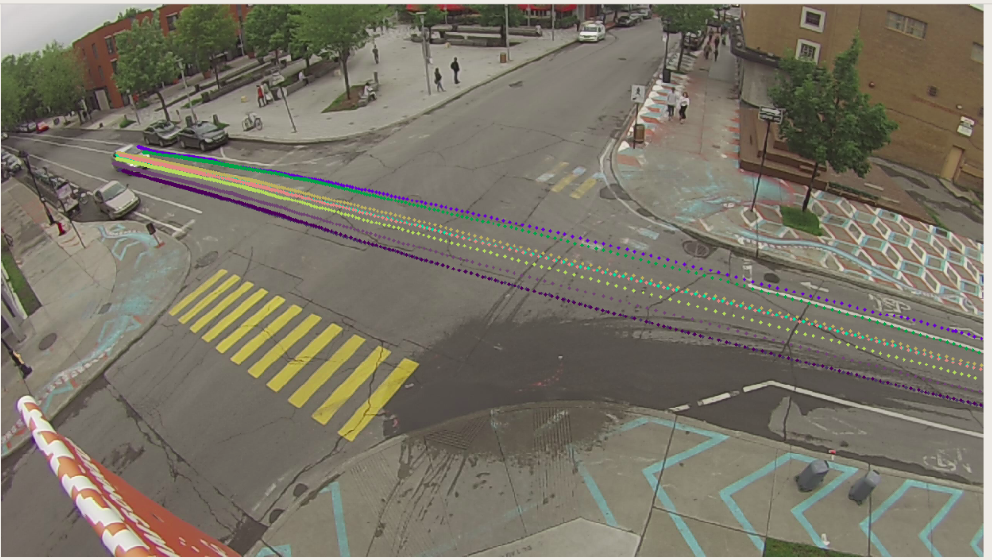}}&
    \subfloat[]{\includegraphics[width=8cm]{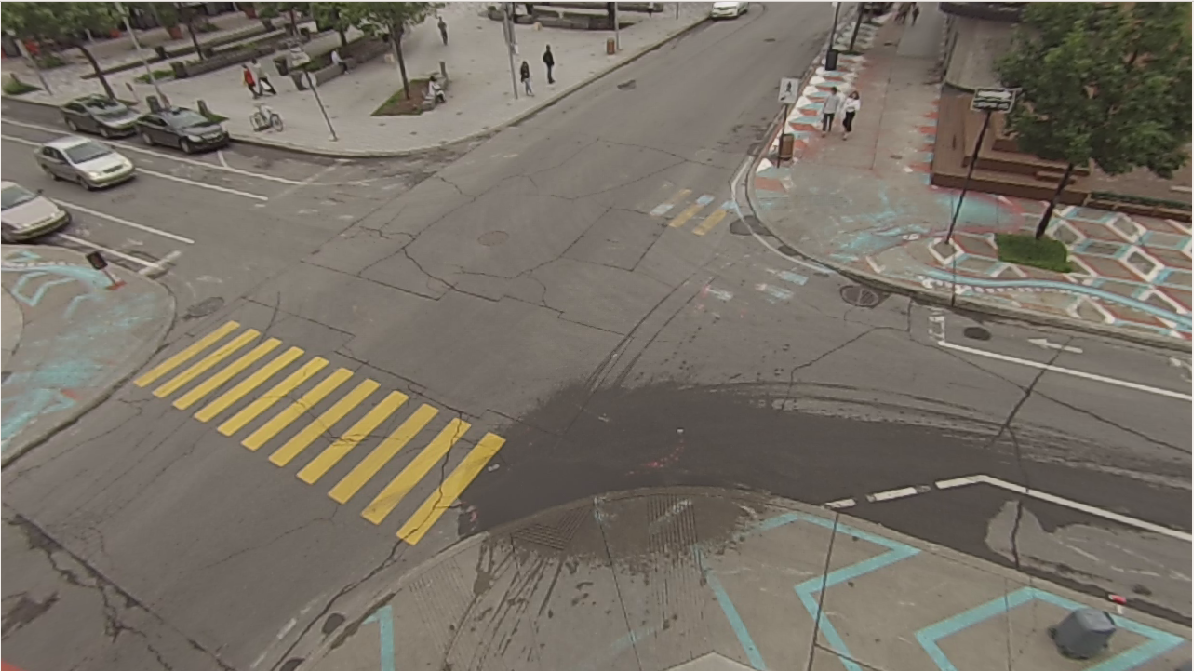}}
\end{tabular}
\caption{(a) Original distorted image of the traffic Scene, (b) Extracting trajectories using KLT-tracker, (c) Filtering top 10 trajectories, (d) Undistorted image of the same scene}%
\label{intrinsic}%

\end{figure*}
\section{Camera Model}
We have assumed our camera to obey the pin-hole camera model. In this model, perspective projection of a 3D point into the 2D image point can be represented as follows :
\begin{equation}{\lambda \begin{bmatrix}u \\ v \\ 1\end{bmatrix} =
\begin{bmatrix}
f_x & s & cx\\
0 & f_y & cy\\
0 & 0 & 1
\end{bmatrix}
\begin{bmatrix}
r_{11} & r_{12} & r_{13} & t_x\\
r_{21} & r_{22} & r_{23} & t_y\\
r_{31} & r_{32} & r_{33} & t_z
\end{bmatrix}
\begin{bmatrix}
X\\
Y\\
Z\\
1
\end{bmatrix}}
\end{equation}
\\The first matrix on R.H.S is referred to as the intrinsic matrix,
as it is only dependent on the internal properties of the camera.
$(f_x, f_y)$ represents the focal length in x and y directions, s
represents the skew present in the pixels. $(cx, cy)$ represents
the optical center. In our approach we have assumed that focal
length in x and y direction are identical, the skew present is zero
and the optical center is also the image center. On simplifying
the equation of a camera which captures an image of size $(H x W)$ becomes:
\begin{equation}
{\lambda \begin{bmatrix}u \\ v \\ 1\end{bmatrix} =
\begin{bmatrix}
f & 0 & W / 2\\
0 & f & H / 2\\
0 & 0 & 1
\end{bmatrix}
\begin{bmatrix}
r_{11} & r_{12} & r_{13} & t_x\\
r_{21} & r_{22} & r_{23} & t_y\\
r_{31} & r_{32} & r_{33} & t_z
\end{bmatrix}
\begin{bmatrix}
X\\
Y\\
Z\\
1
\end{bmatrix}}
\end{equation} 
The second matrix in equation 1 is the extrinsic matrix and
is composed of 3x3 rotation matrix, augmented with a 3x1
translation matrix. In our approach, it would be possible to
estimate the rotation matrices and translation vector (with scale
ambiguity), if no input from the user is provided. However, if
the height of the camera from the ground is also provided, scale
ambiguity could be resolved.
To account for the deviation of the camera from the pin-hole model,
we consider a distortion matrix separately. Here we assume only  radial effects, fitting a polynomial model\cite{Ronda2019} as :
\begin{equation}
    x_u = x_d * (1 + k_1 * r^2 + k_2 * r^4 + k_3 * r^6)
\end{equation}
\begin{equation}
    y_u = y_d * (1 + k_1 * r^2 + k_2 * r^4 + k_3 * r^6)
\end{equation}
Here, $(x_u, y_u), (x_d, y_d)$ denotes the undistorted and distorted normalized coordinates of the image and [k1, k2, k3] are known as distortion coefficients. r is defined as follows:
\begin{equation}
    r_d^2 \xrightarrow{} r^2 = x_d^2 + y_d^2
\end{equation}
In this model, the distortion center coincides with the image center. Also, the fish-eye effect of distortion is assumed to be radial in nature.

\section{Intrinsic Calibration}

\subsection{Fish-eye Effect}\label{AA}
A pin-hole type world to image mapping is possible only by a rectilinear lens which satisfies:
\begin{equation}
    r_u = f * tan(\theta)
\end{equation}
$\theta$  is the angle in in radians between a point in the real world and the optical axis, which goes from the center of the image through the center of the lens. f is the focal length of the lens and $r_u$ is radial position of a point on the image film or sensor. Since it abides by pin-hole model, it is also considered undistorted.

After observation across the datasets available, it was realized, to account for distortion, a global distortion model is to be used. The most common and simple among which was equidistant fish-eye. It is defined as follows :
\begin{equation}
    r_d = f * \theta
\end{equation}
$r_d$ is radial position of a point on the image film or sensor. Since it do not abides by pin-hole model, it is also considered distorted. The equations (6) and (7) can be used for computing forward mapping i.e. from distorted cordinates to undisorted cordinates and inverse mapping i.e. from undistorted cordinates to distorted cordinates respectively as follows :
\begin{equation}
    (x_u, y_u) \xrightarrow{} (x_d * \frac{tan(\frac{r_d}{f})}{\frac{r_d}{f}}, y_d * \frac{tan(\frac{r_d}{f})}{\frac{r_d}{f}} ) 
\end{equation}
\begin{equation}
    (x_d, y_d) \xrightarrow{} (x_u * \frac{atan(\frac{r_u}{f})}{\frac{r_d}{f}} , y_d * \frac{atan(\frac{r_u}{f})}{\frac{r_d}{f}} ) 
\end{equation}
\begin{equation}
    r_u^2 = x_u^2 + y_u^2
\end{equation}
Thus if we have enough correspondences between $(x_d, y_d)$ and $(x_u, y_u)$, it would be possible to estimate the focal length as well as distortion coefficients using a least square approach.

\subsection{Calibration}
To make the calibration independent of the scene, only moving
objects were used. It is done because it is not always possible
to have a similar geometric arrangement of static structures
in every scene on which calibration is performed. However,
moving objects could comprise of pedestrians and vehicles,
either of which is very easily present in every traffic scene.

There is only one unknown in the intrinsic matrix and 
equations 8 and 9, i.e. the focal length f. Thus if we can determine f, the intrinsic matrix, and distortion coefficients
could be computed.
In general, on a road, it is assumed that vehicles move along
a straight line. However, due to distortion(fish-eye effect), these
straight lines become curved in the image. The relation between
the undistorted image and distorted image is only dependent
on the focal length. Thus, by adjusting the value of f to undistort the straight line, focal length can be computed.
The calibration process starts with extraction of moving
object trajectories from video footage as shown in the image.
In a video it is assumed that, there are sufficient number
of vehicles moving in orthogonal directions. To extract the
distorted straight lines, moving pedestrians or vehicles are
tracked in a video footage for a few seconds initially. Tracking
is achieved by optical flow with sift keypoints. Multiple
tracks are extracted and filtered as follows:
\begin{itemize}

\item
If the key point is not tracked for more than 80 \% of the
video interval, it is rejected.
\item
If the total distance in pixels of a keypoints is greater than
1.2 times of its displacement, the keypoints is rejected.
\end{itemize}
Once the trajectories are filtered, top ten longest trajectories are selected. These trajectories are fitted with straight lines
using least square method. The sum of least square errors of all
the trajectories gives us the estimate of the straightness of the
lines. Now the distorted points on trajectories are undistorted by varying the value of focal length from zero to diagonal of
image i.e. the maximum possible focal length in pixels. The
most appropriate focal length would be one with the minimum
least square error. Once the focal length is found, intrinsic
matrix is computed.

However, for estimating the distortion coefficients, the sift keypoints are to be normalized with intrinsic matrices as :
\begin{equation}
    \lambda \begin{bmatrix}x_d \\ y_d \\ 1\end{bmatrix} =
\begin{bmatrix}
f & 0 & W / 2\\
0 & f & H / 2\\
0 & 0 & 1
\end{bmatrix}^{-1}
\begin{bmatrix}
u\\
v\\
1
\end{bmatrix}
\end{equation}
The undistorted cordinates for the same are calculated with the mapping equation 8. Now the cordinates of distorted and undistorted keypoints could be used to compute the Distortion coefficients using equation 11.
\begin{equation}
    \frac{x_u}{x_d} - 1 = \begin{bmatrix}
    r^2 & r^4 & r^6
    \end{bmatrix}\begin{bmatrix}
    k1\\
    k2\\
    k3
    \end{bmatrix}
\end{equation}
The steps of intrinsic calibration are shown in Fig \ref{intrinsic}
\begin{figure}%
    \centering
    \includegraphics[width=8cm]{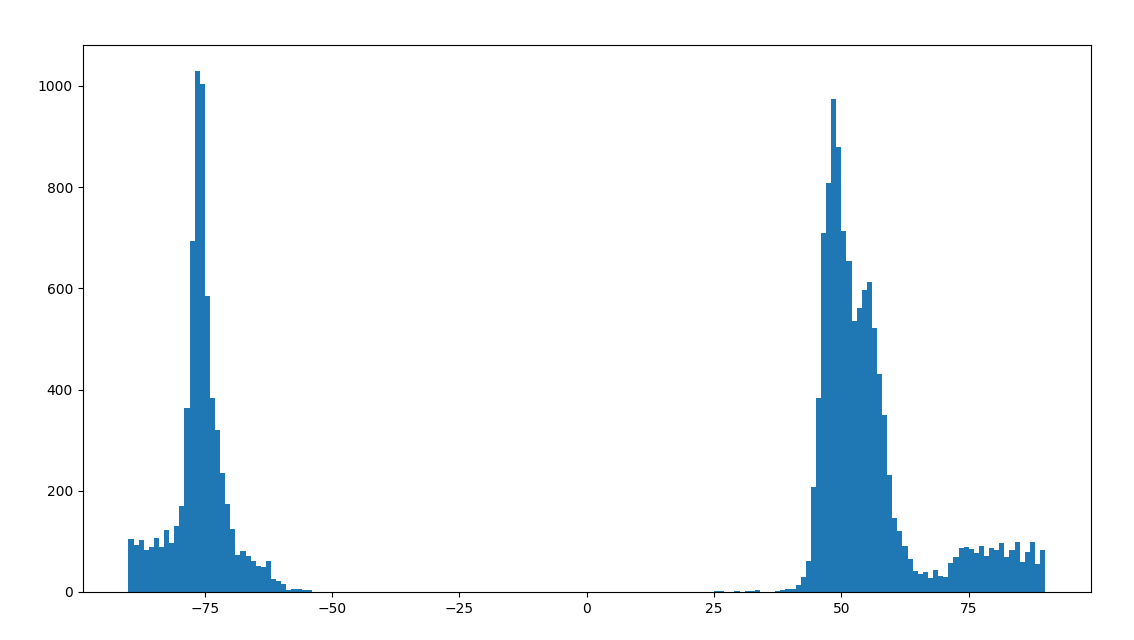}
    \caption{Bimodal distribution of gradients of line segments}%
    \label{histogram}%
\end{figure}
\begin{figure*}%
    \centering
    \subfloat[]{{\includegraphics[width=8cm]{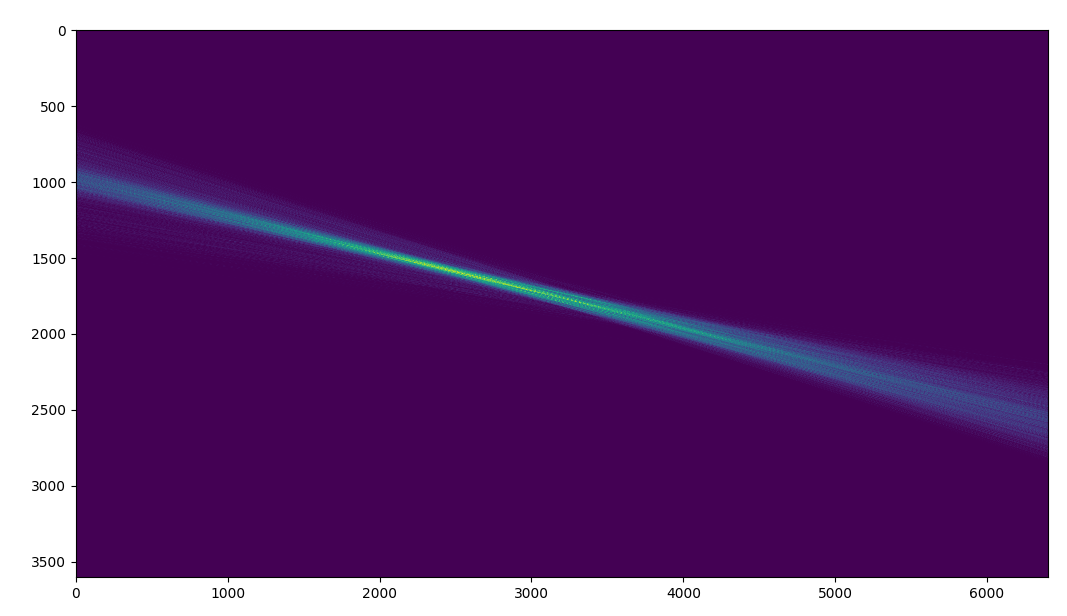} }}%
    \subfloat[]{{\includegraphics[width=8cm]{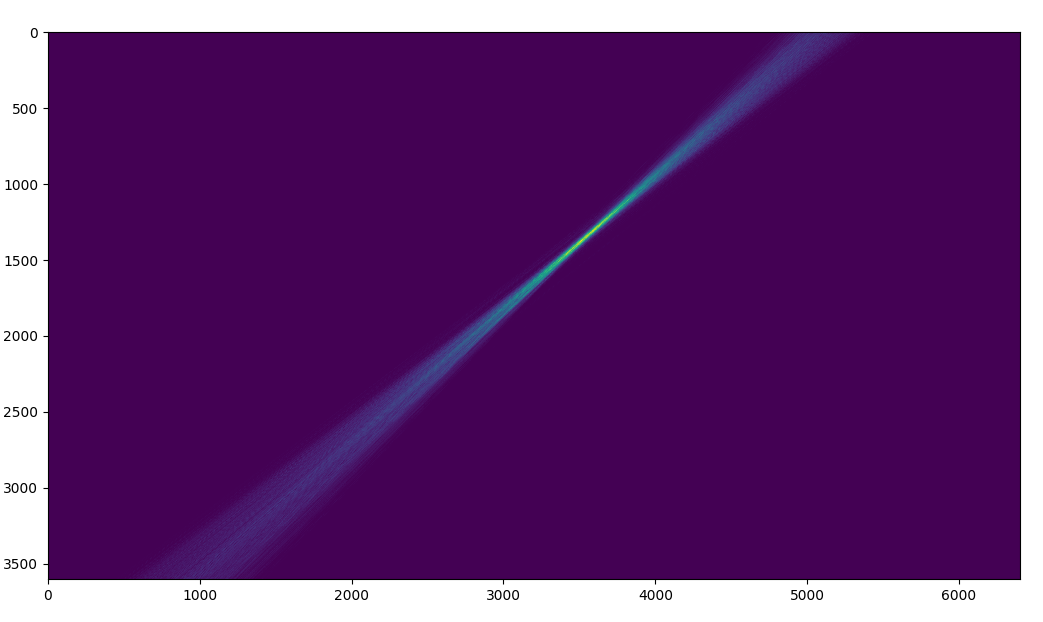} }}%
    \caption{(a) Vanishing Point in x-direction, (b) Vanishing Point in y-direction}%
    \label{votes}%
\end{figure*}
\section{Extrinsic Calibration}
\subsection{Vanishing Point Method}
Given the intrinsics, the extrinsics are computed using vanishing points. Vanishing points are defined as points on image plane where parallel lines in 3D intersect. For calibration, minimum of two orthogonal vanishing points are needed. Due to undistortion, the focal length of the new image is different from the original image. Thus, we will have to first compute the new focal length. If we have two vanishing points say $(u_x, v_x)$ and $(u_y, v_y)$ the new focal length is computed using the relation :
\begin{equation}
    (u_x - c_x)(u_y - c_y) + (v_x - c_y)(v_y - c_y) + f_{new}^2 = 0
\end{equation}
Once the new focal length is known, the elements of rotation matrices computed from equations 14, 15, 16 :
\begin{equation}
    \frac{u_x - c_x}{f_{new}} = \frac{r_{11}}{r_{31}};
    \frac{v_x - c_y}{f_{new}} = \frac{r_{21}}{r_{31}}; r_{11}^2 + r_{21}^2 + r_{31}^2 = 1
\end{equation}
\begin{equation}
    \frac{u_y - c_x}{f_{new}} = \frac{r_{12}}{r_{32}};
    \frac{v_y - c_y}{f_{new}} = \frac{r_{22}}{r_{32}}; r_{12}^2 + r_{22}^2 + r_{32}^2 = 1
\end{equation}
\begin{equation}
    \begin{bmatrix}
    r_{13} \\
    r_{23} \\
    r_{33}
    \end{bmatrix} = \begin{bmatrix}
    r_{11} \\
    r_{21} \\
    r_{31}
    \end{bmatrix} \times \begin{bmatrix}
    r_{12} \\
    r_{22} \\
    r_{32}
    \end{bmatrix} 
\end{equation}

Assuming the road surface to be planar and constitutes a Z = 0 plane, the equation 2 is transformed to : 
\begin{equation}
{\lambda \begin{bmatrix}u \\ v \\ 1\end{bmatrix} =
\begin{bmatrix}
f & 0 & W / 2\\
0 & f & H / 2\\
0 & 0 & 1
\end{bmatrix}
\begin{bmatrix}
r_{11} & r_{12} & t_x\\
r_{21} & r_{22} & t_y\\
r_{31} & r_{32} & t_z
\end{bmatrix}
\begin{bmatrix}
X\\
Y\\
1
\end{bmatrix}}
\end{equation}

If we assume the origin of World Cordinate System corresponds with the center of image and height of the camera from ground is $H_e$ then:
\begin{equation}
    t_x = 0; t_y = 0; t_z = -H_e / r_{33}
\end{equation}

\subsection{Calibration}
As mentioned above, extrinsic calibration is also done using moving objects only. The motion of pedestrians is highly erratic thus, only the motion of vehicles is used for vanishing point estimation. Vehicles in most of the cases follow each other along a straight line. If there are multiple key-points on a single vehicle, it can be seen to originate or converge to a point in image plane, as vehicle moves closer to or farther away from the camera. If there are multiple vehicles moving along two orthogonal directions, orthogonal vanishing points can easily be detected.

The algorithm mentioned below is computationally expensive, and requires significant movement of vehicles in pixels, thus it is performed in one out of every six frames. The first step of calibration is to undistort the image with the help of distortion coefficients, as it is not easy to detect vanishing points from curves. It is followed by YOLO-v3\cite{DBLP:journals/corr/abs-1804-02767} based detection of vehicles. This generates the regions that consists of vehicles with high probability. A mask of such regions is computed for every image on which processing is done. SIFT\cite{Lowe:2004:DIF:993451.996342} key points are generated in the masked images and is matched with the keypoints in the consecutive masked images. The positions of these matched keypoints for any two frames produces multiple line segments. All the line-segments are stored for future processing.

Once, the whole short video-footage is processed, a list of line segments are generated.  Every line segment is converted into polar form i.e (orientation, magnitude, and distance from origin). A histogram of orientation is computed from all the segments as shown in Fig \ref{histogram}. It is observed that it shows a bimodal distribution. The bimodal distribution implies that there are two major directions in which vehicles move. This bimodal distribution helps in generating clusters of line segments in each orthogonal direction. Most appropriate line segments are selectively picked using the distribution peaks with a threshold of 5 degree  in either direction. 

The detection of vanishing point is achieved by voting based system. For each cluster, every line segment is extended to a line. The size of accumulator is equal to one pixel and its value is initially set to zero. Through whichever pixels any line passes, is incremented by one. Thus the pixels with maximum votes are considered most likely positions of vanishing points. The votes in the two directions are shown in the Fig \ref{votes}.

However, the point with maximum votes is not necessarily the vanishing point, and there can be more than one vanishing point  because, we do not achieve a perfect undistortion. If more than one vanishing point exists, it may imply that the point with maximum votes lie some where in between the two or more vanishing points. In order, to account for this, we take the top 20 \% of the pixels with maximum votes and  compute the mean and standard deviation. The estimates of vanishing points is made equal to the sum of mean and two or three times of standard deviation. The columns of rotation matrices does not vary much with the value of constant multiplied with standard deviation.  

Once the vanishing points are computed, the equations 13, 14, 15, 16 and 18, could be used for extrinsic calibration.
To improve the rotation matrix and enforce constraints on it, SVD is performed on the matrix. The new matrix is defined as:
\begin{equation}
    R = USV^T 
\end{equation}
where S is an identity matrix. With the height of camera from the ground, translation vector is computed from equation 18.

\section{Results}
\subsection{Accuracy of Estimated Focal Length}
To estimate the accuracy of the computed focal length from the proposed algorithm, camera is calibrated from checkerboard pattern. When using checkerboard pattern the focal length in x and y direction may not be same, to get the estimate its geometric mean is considered. 

In the table \ref{Focal}, Proposed column is one which uses the proposed algorithm for estimation of focal length and Checkerboard column uses checkerboard calibration. The results from 5 very different video-footages are used to account for generality of results. The accuracy is around 6.35\% of truth value (assumming checkerboard to be ground trruth).
\\
\begin{figure}%
    \centering
    \includegraphics[width=8cm]{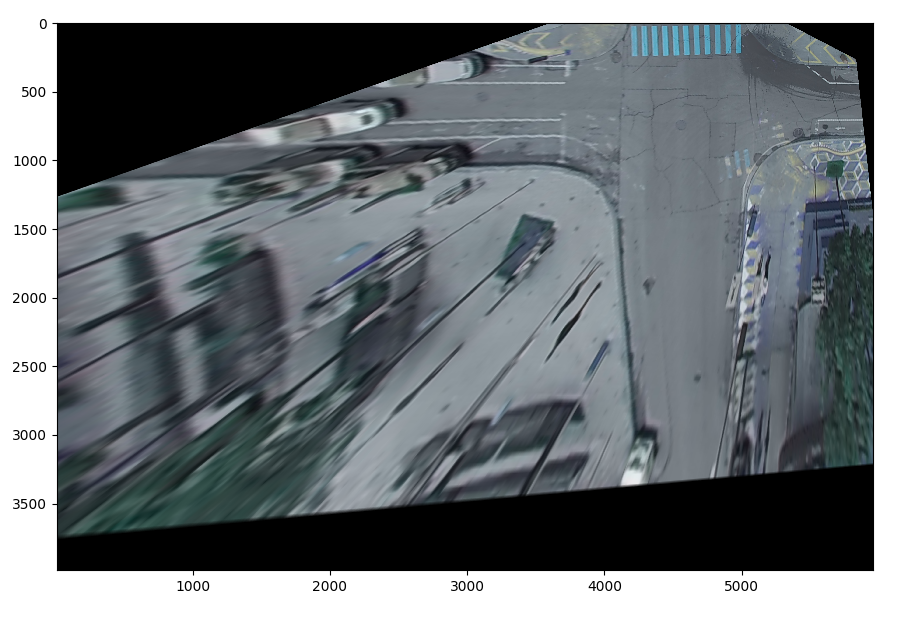}
    \caption{Top-view transformed Image}%
    \label{top-view}%
\end{figure}
\begin{table}
    \begin{tabular}{|c|c|c|}
    \hline
        Datasets & Proposed & Checker Board \\
    \hline
        1. & 1064.20 & 1012.21\\
    \hline
        2. & 619.66 & 584.95\\
    \hline
        3. & 638.52 & 584.95\\
    \hline
        4. & 937.41 & 1012.21\\
    \hline
        5. & 933.43 & 1012.21\\
    \hline
    \end{tabular}
    \captionof{table}{Comparison of focal length with the proposed method and checkerboard method}\label{Focal}
\end{table}
\subsection{Accuracy of Estimated Distortion Coefficients}
To estimate the accuracy of distortion coefficients, mean least square error in trajectories is computed with the predicted distortion coefficients and checkerboard calibrated coefficients. The results are tabulated in Table \ref{Distortion}

In Table \ref{Distortion}, the second column represents the mean error of distortion in trajectory in original image. The third column represents the mean error of distortion in rectified image with proposed algorithm. the last column represents the mean error in distortion in rectified image with checker-board calibration coefficients. It can be seen, in most cases proposed method performs better undistortion then the checkerboard calibrations coefficients.  
\\
\begin{table}[t]
\begin{tabular}{|c|c|c|c|}
    \hline
        Datasets & \pbox{10cm}{Mean LSE \\ before calibration \\for a trajectory} & \pbox{10cm}{Mean LSE\\ after calibration \\for a trajectory} &\pbox{10cm}{Mean LSE after \\checkerboard\\ calibration\\ for a trajectory}\\
    \hline
        1. & 0.572 & 0.103 & 0.149\\
    \hline
        2. & 0.923 & 0.251 & 0.456\\
    \hline
        3. & 0.528 & 0.069 & 0.111\\
    \hline
        4. & 0.166 & 0.004 & 0.061\\
    \hline
        5. & 0.052 & 0.006 & 0.007\\
    \hline
    \end{tabular}

\captionof{table}{Comparison of undistortion of trajectories before and after the Intrinsic Calibration with the proposed method and Checkerboard Method}\label{Distortion}
\end{table}
\subsection{Accuracy of Rotation Matrices}
It was not possible to assess the accuracy of rotation matrices because, the rotation data was not computed when camera was originally set-up, only its video-footage was accessible. However, it is possible to visually estimate its accuracy. It is done by converting the image of a scene to a top-view plane transformed image as shown in Fig \ref{top-view}. In such images, the crosswalks will appear rectangular and not trapezium-like, man-holes will appear circular rather than ellipse, road markings will appear parallel instead of intersecting. There are enough visual cues in such images to test the accuracy of the data visually.

\section {Conclusion and Future Scope}
The proposed algorithm can perform calibration automatically from a video-footage. The intrinsic calibration works well also for the thermal cameras, since calibration procedure is dependent on only moving objects. However, there are certain limitations associated with the proposed method :
\begin{itemize}
    \item Tracking done using KLT tracker is not robust against occlusion.
    \item Assumptions of vehicles move along a straight line may be violated in certain scenarios.
    \item There should be sufficient number of vehicles moving in either directions to estimate vanishing points robust to noise.

Once, the calibration is performed it could be used in multiple applications concerning to photogrammetry, speed-monitoring, 3D-reconstruction, etc. This algorithm in future could also be made independent of the requirement of perpendicular intersections.
\end{itemize}

\end{document}